\documentclass[conference]{IEEEtran}

\usepackage[utf8]{inputenc}
\usepackage[T1]{fontenc}
\usepackage{cite}
\usepackage{amsmath,amsfonts}
\usepackage{graphicx}
\usepackage{booktabs}
\usepackage{multirow}
\usepackage{tabularx}
\usepackage{float}
\usepackage[table]{xcolor}
\usepackage{url}
\usepackage{microtype}
\usepackage{hyperref}
\hypersetup{hidelinks}

\newcolumntype{Y}{>{\centering\arraybackslash}X}
\newcolumntype{Z}{>{\raggedright\arraybackslash}X}

\title{LLM-Enhanced Dual-Branch Learning for Large-Scale Multi-Label Text Classification}

\author{
\IEEEauthorblockN{
Hui Ye\IEEEauthorrefmark{1},
Jing Zhang\IEEEauthorrefmark{2},
Xiulong Yang\IEEEauthorrefmark{3},
Rajshekhar Sunderraman\IEEEauthorrefmark{1}
}
\IEEEauthorblockA{
\IEEEauthorrefmark{1}Department of Computer Science,
Georgia State University, Atlanta, GA 30303, USA\\
\IEEEauthorrefmark{2}Amazon, San Diego, CA, USA\\
\IEEEauthorrefmark{3}School of Computer Science,
Central China Normal University, Wuhan 430079, China\\
hye2@student.gsu.edu, jingz3017@gmail.com,
yangxiulong@ccnu.edu.cn, rsunderraman@gsu.edu
}
}

\begin{document}

\maketitle

\begin{abstract}
Large-scale multi-label text classification assigns a small subset of relevant labels to each document from a vocabulary containing thousands or tens of thousands of candidate labels. Although pretrained language models have improved semantic text representations, most representation-based approaches center their prediction pipelines on a primary encoder or combine auxiliary features within a single ranker. The complementarity between heterogeneous language models therefore remains insufficiently explored. We propose DualMLC, a dual-branch framework that processes the same document through an autoregressive decoder-only language model and a bidirectional encoder. Each branch maintains its own representation pathway and independently estimates relevance scores over the shared label space. DualMLC combines the two score vectors through late logit fusion, allowing shared evidence to reinforce relevant labels and branch-specific evidence to compensate for limitations in the other branch’s representation. DualMLC achieves state-of-the-art results on three widely used large-scale multi-label text classification benchmarks. Ablation results further
confirm that integrating the heterogeneous predictors produces
stronger rankings than either branch alone. The source code is publicly available at https://github.com/huiyegit/DualMLC.

\end{abstract}


\section{Introduction}

Multi-label text classification assigns each document a subset of relevant labels from a predefined vocabulary. In large-scale settings, the vocabulary may contain thousands or tens of thousands of labels, while each document is typically associated with only a small subset. By supporting fine-grained semantic organization and retrieval, the task is valuable in applications including web page tagging~\cite{you2019attentionxml}, product categorization, document tagging, recommendation systems~\cite{zhang2021xrtransformer}, and semantic search.


Research on large-scale multi-label text classification has improved both label prediction efficiency and semantic text representation by recent representative neural systems, including
LightXML~\cite{jiang2021lightxml},
XR-Transformer~\cite{zhang2021xrtransformer},
MatchXML~\cite{ye2024matchxml},
CascadeXML~\cite{kharbanda2022cascadexml}, and
QUEST~\cite{zhou2024quest},
center their prediction pipelines on a primary text encoder. Their label scores are therefore shaped mainly by a single representation geometry. However, bidirectional encoders and autoregressive language models differ substantially in architecture, contextualization mechanism, and pretraining objective, and may therefore capture distinct yet complementary semantic evidence. This distinction is important for prediction over large label vocabularies, where many labels are semantically similar and small changes in text representation can alter their ranking positions. It raises a fundamental scientific question of whether heterogeneous encoders provide genuinely complementary representations and whether their combined evidence can improve the accuracy of the highest ranked labels.

To address this issue, in this paper, we propose a Dual-branch Multi-Label Classification framework, referred to as DualMLC, that explicitly exploits complementary representations for large-scale multi-label text classification. DualMLC processes the same input text through two parallel branches constructed with different representation mechanisms, enabling the input semantics to be modeled from distinct perspectives. Each branch maintains an independent representation pathway and estimates the relevance of candidate labels using its own semantic evidence. This independent learning preserves the distinctive information captured by each representation and avoids prematurely forcing heterogeneous features into a shared space. The prediction evidence produced by the two branches is then combined to form a unified ranking over the label set, allowing shared evidence to reinforce relevant labels and complementary evidence to compensate for limitations of each individual representation. In this way, DualMLC translates representation complementarity into a direct mechanism for improving prediction over large label vocabularies.
Our main contributions are summarized as follows.

\begin{itemize}
    \item We propose DualMLC, a heterogeneous dual branch framework that learns independent text representations through two distinct modeling mechanisms and integrates their prediction evidence for large-scale multi-label text classification.

    \item We investigate representation complementarity over large label vocabularies and empirically demonstrate that the two branches capture distinct yet complementary evidence, with their integration outperforming either branch alone.

    \item Extensive experiments demonstrate that DualMLC achieves new state-of-the-art results on three benchmark datasets, while comprehensive ablation studies verify the contribution of branch integration and the effects of key design choices.
\end{itemize}

\section{Related Work}

Large-scale multi-label learning initially emphasized efficient output modeling. Babbar and Sch\"olkopf~\cite{babbar2017dismec} proposed DiSMEC to retain label specific discrimination with distributed optimization, whereas Yen et al.~\cite{yen2016pdsparse} developed PD-Sparse to reduce cost through primal and dual sparsity. Embedding methods instead compress local label structure. Bhatia et al.~\cite{bhatia2015sleec} learned neighborhood preserving representations in SLEEC, and Tagami~\cite{tagami2017annexml} reconstructed a label neighborhood graph in AnnexML for approximate retrieval. Tree based methods restrict scoring to a candidate path. Jain et al.~\cite{jain2016pfastrexml} introduced propensity scored trees in PfastreXML, Prabhu et al.~\cite{prabhu2018parabel} organized labels with balanced partitions in Parabel, and Khandagale et al.~\cite{khandagale2020bonsai} developed shallower and more diverse trees in Bonsai. Wydmuch et al.~\cite{wydmuch2018extremetext} established a probabilistic foundation for label trees through eXtremeText, while Yu et al.~\cite{yu2022pecos} unified semantic indexing, matching, and sparse ranking in XR-Linear. These methods provide effective mechanisms for controlling label scoring, but their primary concern is output organization rather than heterogeneous semantic text encoding.

Contextual encoders shifted the focus from output structure toward semantic representation learning. Liu et al.~\cite{liu2017deep} proposed XML-CNN to learn convolutional document features, and You et al.~\cite{you2019attentionxml} subsequently combined sequence encoding with label specific attention in AttentionXML. Pretrained Transformers further strengthened this direction. Chang et al.~\cite{chang2020xtransformer} introduced X-Transformer for hierarchical cluster matching, while Jiang et al.~\cite{jiang2021lightxml} developed LightXML to share one Transformer representation between label recall and ranking. Zhang et al.~\cite{zhang2021xrtransformer} extended fine tuning across label resolutions in XR-Transformer, and Kharbanda et al.~\cite{kharbanda2022cascadexml} associated intermediate layers with different resolutions in CascadeXML. A parallel line enriched the prediction signal beyond the primary encoder. Xun et al.~\cite{xun2020cornet} modeled label correlations with CORNet, Dahiya et al.~\cite{dahiya2021siamesexml} aligned instance and label representations in SiameseXML, and Chien et al.~\cite{chien2023pina} incorporated side information through PINA. More recently, Ye et al.~\cite{ye2024matchxml} combined sparse, task tuned, and static sentence representations in MatchXML, while Zhou et al.~\cite{zhou2024quest} adapted a quantized large language model in QUEST. These advances demonstrate the value of richer semantics, but they generally refine one encoder hierarchy or combine multiple feature sources within a single ranking pipeline. DualMLC instead learns architecture distinct encoders as independently supervised predictors and integrates their evidence only after both map to the common label space.

\section{Method}

\subsection{Task Definition}

Let $\mathcal{D}=\{(x_i,\mathbf{y}_i)\}_{i=1}^{N}$ denote a training set of $N$ documents. Each document $x_i$ is associated with a multi-hot vector $\mathbf{y}_i\in\{0,1\}^{L}$ over the label set $\mathcal{L}=\{\ell_1,\ldots,\ell_L\}$. The entry $y_{ij}=1$ indicates that label $\ell_j$ is relevant to $x_i$, while $y_{ij}=0$ indicates otherwise. The task is to learn a scoring function $f_{\theta}(x_i)\in\mathbb{R}^{L}$, parameterized by the model parameter collection $\theta$, that assigns higher scores to relevant labels. For a requested prediction depth $k$, the predicted label set is
\begin{equation}
\widehat{\mathcal{Y}}_i^{(k)}
=
\left\{\ell_j\in\mathcal{L}\mid
j\in\operatorname{TopK}\!\left(f_{\theta}(x_i),k\right)\right\},
\end{equation}
where $\operatorname{TopK}$ returns the indices of the $k$ highest scoring entries. The central modeling requirement is therefore to construct document representations that preserve sufficient semantic evidence for reliable ranking across a large and closely related label set.

\subsection{Overview of DualMLC}

\begin{figure*}[t]
\centering
\includegraphics[width=0.98\textwidth]{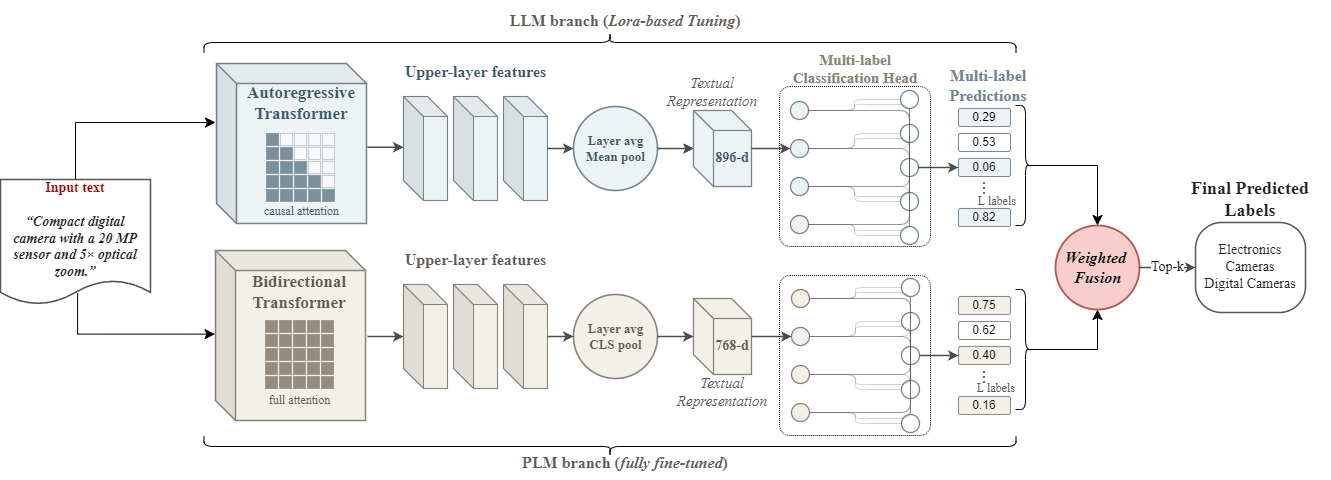}
\vspace{-10pt}
\caption{Overview of DualMLC. The same input text is processed by two heterogeneous representation branches. Each branch independently produces label logits, which are combined to obtain the final label ranking. Both branches receive direct supervision during training.}
\label{fig:architecture}

\vspace{-10pt}
\end{figure*}

DualMLC realizes the scoring function with parallel heterogeneous encoders. We use $q$ for the autoregressive branch and $b$ for the bidirectional branch. For $r\in\{q,b\}$, its logit vector $\mathbf{z}_i^r\in\mathbb{R}^{L}$ is
\begin{equation}
\mathbf{z}_i^r
=
\left(\mathcal{C}_r\circ\mathcal{R}_r\circ\mathcal{P}_r
\circ\mathcal{G}_r\circ\mathcal{E}_r\circ\tau_r\right)(x_i),
\end{equation}
where $\circ$ denotes function composition, $\tau_r$ is the tokenizer, $\mathcal{E}_r$ is the encoder, $\mathcal{G}_r$ aggregates upper layers, $\mathcal{P}_r$ performs sequence pooling, $\mathcal{R}_r$ adjusts dimensionality, and $\mathcal{C}_r$ produces $L$ label logits. Figure~\ref{fig:architecture} summarizes their flow. The branches remain independent until the common label space, which serves as a fusion interface without forcing hidden feature alignment. Each encoder retains its native contextualization and pooling convention, while its classifier maps branch specific evidence to coordinates associated with the same labels.

\subsection{Heterogeneous Text Encoding}

The branch tokenizers $\tau_q$ and $\tau_b$ independently transform a document into two token sequences
\begin{equation}
\mathbf{t}_i^q=\tau_q(x_i),
\qquad
\mathbf{t}_i^b=\tau_b(x_i).
\end{equation}
Separate tokenization respects the different pretrained vocabularies. For document $x_i$ in branch $r\in\{q,b\}$, let $\mathbf{a}_i^r=[a_{i1}^r,\ldots,a_{iT_r}^r]\in\{0,1\}^{T_r}$ denote its binary attention mask of length $T_r$, where $t$ indexes token positions. We set $a_{it}^r=1$ for a nonpadding token and $a_{it}^r=0$ for padding. The mask enters every Transformer layer and excludes padded positions during document pooling. The token states are
\begin{equation}
\begin{aligned}
\mathbf{H}_{i,0}^{r}
&=\operatorname{Emb}_r(\mathbf{t}_i^r),\\
\mathbf{H}_{i,m}^{r}
&=E_{r,m}\!\left(\mathbf{H}_{i,m-1}^{r},\mathbf{a}_i^r\right),
\quad m=1,\ldots,M_r,
\end{aligned}
\end{equation}
where $\operatorname{Emb}_r$ and $E_{r,m}$ are the embedding layer and the $m$th Transformer layer. The state $\mathbf{H}_{i,m}^{r}\in\mathbb{R}^{T_r\times d_r}$ has hidden width $d_r$, and $M_r$ is the number of layers. Branch $q$ uses causal attention, whereas branch $b$ uses bidirectional attention. Causal attention summarizes a document under an autoregressive constraint, while bidirectional attention integrates evidence from both sides of every token. Their different context formation mechanisms can therefore emphasize different lexical and semantic cues for the same label, providing the representational diversity required by the dual branch design.

Branch $q$ uses Qwen2.5-7B~\cite{qwen2024qwen25}, while branch $b$ uses BERT-base-uncased~\cite{devlin2019bert}. We adapt the larger autoregressive backbone with low rank adaptation~\cite{hu2022lora} in each attention projection indexed by $p\in\{Q,K,V,O\}$, where $Q$, $K$, $V$, and $O$ denote the query, key, value, and output projections. For input $\mathbf{u}\in\mathbb{R}^{d_{\mathrm{in}}}$ at layer $m\in\{1,\ldots,M_q\}$, the adapted projection is
\begin{equation}
\operatorname{Proj}_{m,p}(\mathbf{u})
=
\mathbf{W}_{m,p}^{0}\mathbf{u}
+\frac{\gamma_{\mathrm{L}}}{r_{\mathrm{L}}}
\mathbf{B}_{m,p}\mathbf{A}_{m,p}
\mathcal{D}_{\mathrm{L}}(\mathbf{u}),
\end{equation}
where $d_{\mathrm{in}}$ and $d_{\mathrm{out}}$ are the input and output widths of the projection. The frozen pretrained weight is $\mathbf{W}_{m,p}^{0}\in\mathbb{R}^{d_{\mathrm{out}}\times d_{\mathrm{in}}}$, whose superscript $0$ distinguishes it from the learned update. The factors $\mathbf{A}_{m,p}\in\mathbb{R}^{r_{\mathrm{L}}\times d_{\mathrm{in}}}$ and $\mathbf{B}_{m,p}\in\mathbb{R}^{d_{\mathrm{out}}\times r_{\mathrm{L}}}$ are trainable. Moreover, $r_{\mathrm{L}}$ is the LoRA rank, $\gamma_{\mathrm{L}}$ is the scaling factor, and $\mathcal{D}_{\mathrm{L}}$ is the LoRA dropout operator applied to $\mathbf{u}$. The product $\mathbf{B}_{m,p}\mathbf{A}_{m,p}$ constitutes a task specific update while leaving $\mathbf{W}_{m,p}^{0}$ unchanged. Its rank controls the adaptation subspace, while $\gamma_{\mathrm{L}}/r_{\mathrm{L}}$ controls the update scale. This design limits the trainable parameters of branch $q$, whereas branch $b$ is fully fine tuned to retain task specific flexibility.

\subsection{Layer Aggregation and Document Representation}

Because useful semantics may remain distributed across upper layers, we average the last $s_r$ layers before pooling. Uniform averaging introduces no trainable parameters and reduces dependence on a single final layer. For a branch with $M_r$ layers, the aggregated states are
\begin{equation}
\overline{\mathbf{H}}_i^r
=
\frac{1}{s_r}
\sum_{j=0}^{s_r-1}
\mathbf{H}_{i,M_r-j}^{r}.
\end{equation}
Pooling follows the encoder conventions. With $\overline{\mathbf{H}}_{i,t}^{r}$ denoting the $t$th token state, branch $q$ uses masked mean pooling and branch $b$ uses the first classification token
\begin{align}
\mathbf{h}_i^q
&=
\frac{\sum_{t=1}^{T_q}a_{it}^{q}\overline{\mathbf{H}}_{i,t}^{q}}
{\sum_{t=1}^{T_q}a_{it}^{q}},
\\
\mathbf{h}_i^b
&=
\overline{\mathbf{H}}_{i,1}^{b}.
\end{align}
This avoids imposing the same sequence summary on different encoder architectures.

Since $d_q$ is substantially larger than $d_b$, we use parameter free grouped mean reduction to control the label head size. For a group size $g$ that divides $d_q$, the $j$th component of $\widetilde{\mathbf{h}}_i^q\in\mathbb{R}^{d_q/g}$ is
\begin{equation}
\left[\widetilde{\mathbf{h}}_i^q\right]_j
=
\frac{1}{g}
\sum_{s=1}^{g}
\left[\mathbf{h}_i^q\right]_{g(j-1)+s},
\qquad j=1,\ldots,d_q/g.
\end{equation}
No reduction is applied to branch $b$, so $\widetilde{\mathbf{h}}_i^b=\mathbf{h}_i^b$. The grouped operation reduces the branch $q$ classifier from $Ld_q$ to $Ld_q/g$ weights while preserving feature scale. It does not require the two representations to have the same dimension because comparability is established only after both are mapped to $L$ label logits.

\subsection{Branch Specific Label Prediction}

Let $\widetilde d_q=d_q/g$ and $\widetilde d_b=d_b$ denote the two classifier input dimensions. Each branch uses an independent linear classifier to map its document representation into the common label space
\begin{equation}
\mathbf{z}_i^r
=
\mathbf{W}_r\operatorname{Dropout}\!\left(\widetilde{\mathbf{h}}_i^r\right)
+\mathbf{c}_r,
\qquad r\in\{q,b\},
\end{equation}
where $\mathbf{W}_r\in\mathbb{R}^{L\times\widetilde d_r}$ and $\mathbf{c}_r\in\mathbb{R}^{L}$ are classifier parameters, and $\operatorname{Dropout}$ is classifier dropout. Thus each branch learns its own semantic mapping, while the shared label coordinates make the logit vectors directly comparable.

\begin{table*}[!t]
\centering
\caption{P@$k$ (\%) on three multi-label text classification benchmarks. The best and second-best results are shown in bold and underlined, respectively.}
\vspace{-5pt}
\label{tab:main_results}
\footnotesize
\setlength{\tabcolsep}{3pt}
\renewcommand{\arraystretch}{1.08}
\begin{tabularx}{\textwidth}{l*{9}{Y}}
\toprule
\multirow{2}{*}{Method}
& \multicolumn{3}{c}{EURLex-4K}
& \multicolumn{3}{c}{Wiki10-31K}
& \multicolumn{3}{c}{AmazonCat-13K} \\
\cmidrule(lr){2-4}\cmidrule(lr){5-7}\cmidrule(lr){8-10}
& P@1 & P@3 & P@5
& P@1 & P@3 & P@5
& P@1 & P@3 & P@5 \\
\midrule
AnnexML~\cite{tagami2017annexml}
& 79.66 & 69.64 & 53.52
& 86.46 & 74.28 & 64.20
& 93.54 & 78.36 & 63.30 \\
DiSMEC~\cite{babbar2017dismec}
& 83.21 & 70.39 & 58.73
& 84.13 & 74.72 & 65.94
& 93.81 & 79.08 & 64.06 \\
PfastreXML~\cite{jain2016pfastrexml}
& 73.14 & 60.16 & 50.54
& 83.57 & 68.61 & 59.10
& 91.75 & 77.97 & 63.68 \\
Parabel~\cite{prabhu2018parabel}
& 82.12 & 68.91 & 57.89
& 84.19 & 72.46 & 63.37
& 93.02 & 79.14 & 64.51 \\
eXtremeText~\cite{wydmuch2018extremetext}
& 79.17 & 66.80 & 56.09
& 83.66 & 73.28 & 64.51
& 92.50 & 78.12 & 63.51 \\
Bonsai~\cite{khandagale2020bonsai}
& 82.30 & 69.55 & 58.35
& 84.52 & 73.76 & 64.69
& 92.98 & 79.13 & 64.46 \\
XR-Linear~\cite{yu2022pecos}
& 84.14 & 72.05 & 60.67
& 85.75 & 75.79 & 66.69
& 94.64 & 79.98 & 64.79 \\
\midrule
XML-CNN~\cite{liu2017deep}
& 75.32 & 60.14 & 49.21
& 81.41 & 66.23 & 56.11
& 93.26 & 77.06 & 61.40 \\
AttentionXML~\cite{you2019attentionxml}
& 85.49 & 73.08 & 61.10
& 87.10 & 77.80 & 68.80
& 95.65 & 81.93 & 66.90 \\
LightXML~\cite{jiang2021lightxml}
& 86.02 & 74.02 & 61.87
& 87.80 & 77.30 & 68.00
& \underline{96.55} & \underline{83.70} & \underline{68.46} \\
APLC-XLNet~\cite{ye2020aplc}
& 83.60 & 70.20 & 57.90
& 88.76 & 79.11 & 69.63
& 96.14 & 82.86 & 67.58 \\
XR-Transformer~\cite{zhang2021xrtransformer}
& 87.22 & 74.39 & 61.69
& 88.00 & 78.70 & 69.10
& 96.25 & 82.72 & 67.01 \\
MatchXML~\cite{ye2024matchxml}
& \underline{88.12} & \underline{75.00} & \underline{62.22}
& \underline{89.30} & \underline{80.45} & \underline{70.89}
& 96.50 & 83.25 & 67.69 \\
\midrule
\rowcolor{gray!10}
\textbf{DualMLC (Ours)}
& \textbf{88.82} & \textbf{75.80} & \textbf{62.74}
& \textbf{90.78} & \textbf{81.11} & \textbf{71.94}
& \textbf{96.90} & \textbf{84.39} & \textbf{69.14} \\
\bottomrule
\end{tabularx}
\vspace{-10pt}
\end{table*}

\subsection{Logit Fusion and Inference}

The common output space enables late fusion according to
\begin{equation}
\mathbf{z}_i
=
\alpha\mathbf{z}_i^q
+(1-\alpha)\mathbf{z}_i^b,
\qquad 0\leq\alpha\leq1,
\end{equation}
where $\mathbf{z}_i\in\mathbb{R}^{L}$ is the fused logit vector and $\alpha$ controls the contribution of branch $q$. This preserves branch independence while allowing complementary evidence to compensate for a weak branch score. The probabilities are $\widehat{\mathbf{p}}_i=\sigma(\mathbf{z}_i)$, where $\sigma$ is the elementwise sigmoid. Its monotonicity preserves the logit ranking, so inference returns the labels with the $k$ largest entries of $\mathbf{z}_i$.

\subsection{Training Objective}

Both branches receive direct supervision from the same multi-hot target. For branch $r\in\{q,b\}$, with $z_{ij}^{r}$ denoting the $j$th entry of $\mathbf{z}_i^r$, the binary cross entropy is
\begin{equation}
\begin{split}
\mathcal{L}_{r}
=
-\frac{1}{NL}
\sum_{i=1}^{N}\sum_{j=1}^{L}
\bigl[
&y_{ij}\log\sigma(z_{ij}^{r})
\\
&+(1-y_{ij})\log\!\left(1-\sigma(z_{ij}^{r})\right)
\bigr].
\end{split}
\end{equation}
Let $\Theta_q$ and $\Theta_b$ be the trainable blocks of $\theta$. The former contains the LoRA factors and branch $q$ classifier, while the latter contains the branch $b$ encoder and classifier. Training solves
\begin{equation}
\left(\Theta_q^{\star},\Theta_b^{\star}\right)
=
\underset{\Theta_q,\Theta_b}{\arg\min}
\left[\mathcal{L}_{q}(\Theta_q)+\mathcal{L}_{b}(\Theta_b)\right].
\end{equation}
The superscript $\star$ denotes the optimized parameter values. The frozen autoregressive backbone and fixed $\alpha$ are excluded from these sets, and no loss is applied to the fused logits. Separate losses preserve the informativeness of each predictor, allowing their logits to be evaluated separately and combined only during late fusion.

\begin{table}[!ht]
\centering
\caption{Statistics of the benchmark datasets.}
\vspace{-5pt}
\label{tab:dataset_statistics}
\renewcommand{\arraystretch}{1.05}
\setlength{\tabcolsep}{3.0pt}
\resizebox{\columnwidth}{!}{%
\begin{tabular}{lrrrrr}
\toprule
Dataset & Train & Test & Labels & Avg. Labels & Avg. Instances \\
\midrule
EURLex-4K     & 15,449    & 3,865   & 3,956  & 5.30  & 20.79 \\
Wiki10-31K    & 14,146    & 6,616   & 30,938 & 18.64 & 8.52 \\
AmazonCat-13K & 1,186,239 & 306,782 & 13,330 & 5.04  & 448.57 \\
\bottomrule
\end{tabular}%
}
\vspace{-10pt}
\end{table}

\section{Experiments}

\subsection{Experimental Settings}

\subsubsection{Datasets}

We evaluate DualMLC on three widely used multi-label text classification benchmarks that differ in domain, label cardinality, and training scale. EURLex-4K~\cite{loza2010eurlex} contains European Union legal documents annotated with EuroVoc concepts. Wiki10-31K~\cite{zubiaga2009wiki10} consists of English Wikipedia articles paired with social tags collected from Delicious. AmazonCat-13K~\cite{mcauley2013amazon} contains Amazon product text associated with multiple product categories. Table~\ref{tab:dataset_statistics} summarizes the benchmark splits used in our experiments.

Following prior work, we report precision at one, three, and five, denoted by P@1, P@3, and P@5. In general, P@$k$ is the fraction of the top $k$ predictions that belong to the ground truth label set, averaged over all test instances. 

\subsubsection{Baselines}

We compare DualMLC with thirteen representative methods and organize them into two groups. Sparse, embedding, and label tree methods include AnnexML~\cite{tagami2017annexml}, DiSMEC~\cite{babbar2017dismec}, PfastreXML~\cite{jain2016pfastrexml}, Parabel~\cite{prabhu2018parabel}, eXtremeText~\cite{wydmuch2018extremetext}, Bonsai~\cite{khandagale2020bonsai}, and XR-Linear~\cite{yu2022pecos}, covering local output embeddings, sparse label classifiers, and hierarchical routing. Neural text encoder methods include XML-CNN~\cite{liu2017deep}, AttentionXML~\cite{you2019attentionxml}, LightXML~\cite{jiang2021lightxml}, APLC-XLNet~\cite{ye2020aplc}, XR-Transformer~\cite{zhang2021xrtransformer}, and MatchXML~\cite{ye2024matchxml}, covering convolutional, recurrent, and pretrained Transformer encoders.

\subsubsection{Implementation Details}

By default, both branches truncate or pad inputs to 256 tokens. We average the last four hidden layers of both encoders, apply attention mask aware mean pooling to Qwen2.5-7B, and use the classification token, denoted by CLS, from BERT-base-uncased. A dropout rate of 0.1 is applied before each linear classifier. The 3,584 dimensional Qwen representation is reduced to 896 dimensions, while the 768 dimensional BERT representation is used directly.

The Qwen backbone remains frozen and is adapted with LoRA on the query, key, value, and output projections of every attention block. We set the LoRA rank to 16, its scaling parameter to 32, and its dropout rate to 0.05. BERT is fine tuned end to end. We optimize the sum of the two branch losses using AdamW with a weight decay of 0.01. The learning rates for the Qwen LoRA parameters, Qwen classifier, BERT encoder, and BERT classifier are $5\times10^{-5}$, $5\times10^{-4}$, $1\times10^{-4}$, and $2\times10^{-3}$, respectively. The gradients are clipped to a norm of 1.0. The training and evaluation batch sizes are 2 and 32 per device. Training uses bfloat16 mixed precision and a random seed of 42. We set $\alpha=0.6$, which assigns weights of 0.6 and 0.4 to the Qwen and BERT logits. All experiments are conducted on a server equipped with eight NVIDIA GeForce RTX 4090 GPUs with 24 GB of memory each.

\subsection{Main Results}

Table~\ref{tab:main_results} shows that DualMLC achieves the highest value for all nine dataset and metric combinations. On EURLex-4K, it improves over MatchXML by 0.70\%, 0.80\%, and 0.52\% in P@1, P@3, and P@5. The corresponding gains over MatchXML on Wiki10-31K are 1.48\%, 0.66\%, and 1.05\%. On AmazonCat-13K, where LightXML is the strongest baseline for all three metrics, DualMLC improves P@1, P@3, and P@5 by 0.35\%, 0.69\%, and 0.68\%.

The gains are consistent across benchmarks with different label vocabularies and training scales, which indicates that the benefit is not confined to one data regime. The distribution of improvements across ranking depths is also informative. EURLex-4K shows relatively even gains from P@1 to P@5, whereas Wiki10-31K obtains its largest improvement at P@1. This result is notable because Wiki10-31K has the largest label vocabulary among the evaluated datasets, containing 30,938 labels. On AmazonCat-13K, the gains at P@3 and P@5 exceed the gain at P@1, indicating that fusion remains useful as more labels are retained. 

\vspace{-5pt}
\begin{table}[H]
\centering
\caption{Comparison of prediction branches on Wiki10-31K.}
\vspace{-5pt}
\label{tab:ablation_branch}
\footnotesize
\renewcommand{\arraystretch}{1.06}
\begin{tabularx}{0.90\columnwidth}{ZYYY}
\toprule
Branch & P@1 & P@3 & P@5 \\
\midrule
BERT & 88.74 & 77.99 & 68.13 \\
Qwen & 90.02 & 79.81 & 70.94 \\
\rowcolor{gray!25}
DualMLC & \textbf{90.78} & \textbf{81.11} & \textbf{71.94} \\
\bottomrule
\end{tabularx}
\vspace{-5pt}
\end{table}
\vspace{-12pt}

\begin{table}[!ht]
\centering
\caption{Effect of the fusion weight on Wiki10-31K.}
\vspace{-5pt}
\label{tab:ablation_alpha}
\footnotesize
\renewcommand{\arraystretch}{1.06}
\begin{tabularx}{0.90\columnwidth}{YYYY}
\toprule
$\alpha$ & P@1 & P@3 & P@5 \\
\midrule
0.5 & \textbf{90.80} & 81.00 & 71.91 \\
\rowcolor{gray!25}
0.6 & 90.78 & \textbf{81.11} & \textbf{71.94} \\
0.7 & 90.25 & 80.59 & 71.68 \\
0.8 & 89.75 & 80.39 & 71.53 \\
\bottomrule
\end{tabularx}
\vspace{-5pt}
\end{table}

\subsection{Ablation Study}

\textit{Branch complementarity.} Table~\ref{tab:ablation_branch} shows that Qwen is stronger than BERT when each branch is evaluated alone, with gains of 1.28\%, 1.82\%, and 2.81\% in P@1, P@3, and P@5. DualMLC  improves over Qwen by a further 0.76\%, 1.30\%, and 1.00\%. The weaker individual branch therefore contributes label evidence that is not recovered by selecting only the stronger encoder. Moreover, the fusion gains of   P@3 and P@5, indicate that the contribution of BERT is not limited to correcting the first prediction. It also improves the ordering of additional relevant labels.

\subsection{Hyper-parameter Analysis}

\textit{Fusion and pooling.} Table~\ref{tab:ablation_alpha} shows that equal weighting gives the highest P@1 of 90.80\%, while $\alpha=0.6$ gives the highest P@3 and P@5 of 81.11\% and 71.94\%. The difference between $\alpha=0.5$ and $\alpha=0.6$ remains small, whereas performance declines more clearly as the Qwen weight increases to 0.8. The method is therefore stable near balanced fusion but benefits from retaining meaningful evidence from both predictors. This behavior motivates $\alpha=0.6$ as a balanced setting. With mean pooling fixed for Qwen, Table~\ref{tab:ablation_pooling} shows that replacing BERT mean pooling with its classification token representation improves P@1, P@3, and P@5 by 1.04\%, 0.79\%, and 0.94\%.

\begin{table}[t]
\centering
\caption{Effect of pooling strategies on Wiki10-31K.}
\vspace{-5pt}
\label{tab:ablation_pooling}
\footnotesize
\renewcommand{\arraystretch}{1.06}
\begin{tabularx}{0.90\columnwidth}{ZZYYY}
\toprule
Qwen Pool & BERT Pool & P@1 & P@3 & P@5 \\
\midrule
\rowcolor{gray!25}
Mean & CLS & \textbf{90.78} & \textbf{81.11} & \textbf{71.94} \\
Mean & Mean & 89.74 & 80.32 & 71.00 \\
\bottomrule
\end{tabularx}
\vspace{-5pt}
\end{table}

\begin{table}[t]
\centering
\caption{Effect of the number of averaged upper layers on Wiki10-31K.}
\vspace{-5pt}
\label{tab:ablation_layers}
\footnotesize
\renewcommand{\arraystretch}{1.06}
\begin{tabularx}{0.90\columnwidth}{YYYYY}
\toprule
$s_q$ & $s_b$ & P@1 & P@3 & P@5 \\
\midrule
1 & 1 & 90.40 & 80.80 & 71.36 \\
1 & 4 & 90.36 & 80.98 & 71.68 \\
2 & 2 & 90.39 & 80.89 & \textbf{71.97} \\
\rowcolor{gray!25}
4 & 4 & \textbf{90.78} & \textbf{81.11} & 71.94 \\
6 & 6 & 90.66 & 80.77 & 71.83 \\
\bottomrule
\end{tabularx}
\vspace{-5pt}
\end{table}

\begin{table}[t]
\centering
\caption{Effect of input sequence length on Wiki10-31K.}
\vspace{-5pt}
\label{tab:ablation_length}
\footnotesize
\renewcommand{\arraystretch}{1.06}
\begin{tabularx}{0.90\columnwidth}{YYYYY}
\toprule
$T_q$ & $T_b$ & P@1 & P@3 & P@5 \\
\midrule
128 & 128 & 90.31 & 80.48 & 71.43 \\
\rowcolor{gray!25}
256 & 256 & \textbf{90.78} & \textbf{81.11} & 71.94 \\
512 & 512 & 90.36 & 81.04 & \textbf{72.03} \\
\bottomrule
\end{tabularx}
\vspace{-8pt}
\end{table}

\begin{table}[t]
\centering
\caption{Effect of the LoRA rank setting on Wiki10-31K.}
\vspace{-5pt}
\label{tab:ablation_lora}
\footnotesize
\renewcommand{\arraystretch}{1.06}
\begin{tabularx}{0.90\columnwidth}{YYYY}
\toprule
$r_{\mathrm{L}}$ & P@1 & P@3 & P@5 \\
\midrule
4  & 90.66 & 81.04 & 71.94 \\
8  & 90.63 & 80.78 & \textbf{71.97} \\
\rowcolor{gray!25}
16 & \textbf{90.78} & \textbf{81.11} & 71.94 \\
32 & \textbf{90.78} & 81.04 & 71.94 \\
\bottomrule
\end{tabularx}
\vspace{-10pt}
\end{table}

\textit{Layer aggregation and input length.} Table~\ref{tab:ablation_layers} shows that averaging the last four layers of both encoders improves P@1, P@3, and P@5 over using only the final layer by 0.38\%, 0.31\%, and 0.58\%. The four layer setting gives the best P@1 and P@3, while averaging two layers gives a marginally higher P@5 by 0.03\%. Extending the average to six layers produces no further gain. Table~\ref{tab:ablation_length} shows that a sequence length of 256 provides the best balance across ranking depths. Compared with length 128, it improves the three metrics by 0.47\%, 0.63\%, and 0.51\%. Increasing the length to 512 raises P@5 by only 0.09\% and lowers P@1 and P@3, so the added computation does not yield a consistent accuracy benefit. 
Upper layer averaging combines several semantic abstractions, whereas including too many layers can mix features that are less aligned with the downstream labels. Similarly, a longer sequence preserves more text but raises attention cost and may introduce weakly relevant tokens. The selected four layer and 256 token configuration therefore balances semantic coverage, ranking stability, and computational efficiency.

\textit{LoRA configuration.} Table~\ref{tab:ablation_lora} shows modest variation across the evaluated rank settings. Rank 16 gives the best P@3 and ties for the best P@1, while rank 8 exceeds it in P@5 by only 0.03\%. Across ranks from 4 to 32, P@1 varies by 0.15\%, P@3 by 0.33\%, and P@5 by 0.03\%. Since $\gamma_{\mathrm{L}}$ remains fixed, changing $r_{\mathrm{L}}$ also changes the effective scale $\gamma_{\mathrm{L}}/r_{\mathrm{L}}$. This table therefore compares practical LoRA settings rather than isolating rank capacity alone. The limited sensitivity indicates that DualMLC does not depend on a narrowly tuned adaptation size. Rank 16 is preferred because it jointly attains the strongest P@1 and P@3 without increasing the trainable parameters to the rank 32 setting.

\subsection{Discussion}

The ablation and sensitivity results indicate that DualMLC benefits from complementary representations rather than capacity alone. Causal contextualization emphasizes progressively formed context and generative pretraining knowledge, while bidirectional contextualization evaluates tokens using evidence from both directions. Independent supervision preserves these distinct cues before fusion in a common label space.

\section{Conclusion}

This work investigated whether heterogeneous language models provide complementary evidence for large-scale multi-label text classification. We introduced DualMLC, which preserves independent representation and prediction pathways and integrates their label scores through late logit fusion. Experiments on three benchmarks have showed consistent improvements across datasets and ranking depths. Branch ablations further showed that the combined predictor outperforms either branch alone, supporting the view that heterogeneous representations capture useful nonredundant evidence. These findings establish representation complementarity as a promising direction for prediction over large label vocabularies. Future work will integrate DualMLC with hierarchical or clustered label indexing to improve efficiency on substantially larger label spaces.

\section*{Acknowledgments}

Research was sponsored by the Army Research Laboratory and was accomplished under Cooperative Agreement Number W911NF-23-2-0224. The views and conclusions contained in this document are those of the authors and should not be interpreted as representing the official policies, either expressed or implied, of the Army Research Laboratory or the U.S. Government. The U.S. Government is authorized to reproduce and distribute reprints for Government purposes notwithstanding any copyright notation herein.

\end{document}